%% file: ms.tex
\pdfoutput=1
\def\year{2018}\relax
\documentclass[letterpaper]{article} %DO NOT CHANGE THIS
\usepackage{aaai18}  %Required
\usepackage{times}  %Required
\usepackage{helvet}  %Required
\usepackage{courier}  %Required
\usepackage{url}  %Required
\usepackage{graphicx}  %Required
\usepackage{booktabs}       % professional-quality tables
\usepackage[cmex10]{amsmath}
\usepackage{bm}
\usepackage{amssymb}

\usepackage{algorithm}
\usepackage[noend]{algpseudocode}

\usepackage{tikz}
\usepackage{xcolor}
\usepackage{makeidx}
\usepackage{graphicx}
\usepackage{multirow}
\usepackage{hhline}

\usepackage{wrapfig}
\usepackage{graphicx}
\usepackage{caption}
\usepackage{subcaption}
\graphicspath{ {data-plots/binned-cross-plots/} }
\usepackage{mathtools}

\usepackage{float}
\usepackage{titlesec}
\usepackage{lipsum}

\newtheorem{example}{Example}
\newtheorem{definition}{Definition}

\makeatletter
\def\@BracContents{} % default (used by \suchthat)
\newcommand{\BracKern}{\kern-\nulldelimiterspace}
\newcommand{\@Brac}[3]{% #1,#3 = left/right bracket type
	\ensuremath{%
		\begingroup\def\@BracContents{#2}%
		\mathopen{\left#1\vphantom{#2}\BracKern\right.}% left bracket
		#2%  content
		\mathclose{\left.\BracKern\vphantom{#2}\right#3}% right bracket
		\endgroup%
	}%
}
\newcommand{\bracr}[1]{\@Brac{(}{#1}{)}}%
\makeatother

\selectcolormodel{gray}

\newcommand{\ourModel}{\textsf{JoinInfer}}
\newcommand{\ourModelDown}{\textsf{JoinInferDown}}
\newcommand{\ourModelUp}{\textsf{JoinInferUp}}
\newcommand{\ourModelSampling}{\textsf{JoinInferSampling}}

\newcommand{\NPRR}{\textsf{MultFacProd}}
\newcommand{\parent}{\textsf{Parent}}
\newcommand{\children}{\textsf{Children}}
\newcommand{\bandone}{Band $1$}
\newcommand{\bandtwo}{Band $2$}
\newcommand{\bandthree}{Band $3$}
\newcommand{\bandfour}{Band $4$}
\newcommand{\bandfive}{Band $5$}
\newcommand{\bandsix}{Band $6$}
\newcommand{\PP}{\textsf{PairwiseProd}}

\newcommand{\tw}{\mathrm{tw}}

\newcommand{\fhtw}{\mathrm{fhtw}}
\newcommand{\htw}{\mathrm{htw}}

\input{title_and_authors}
\begin{document}
\maketitle
%\onecolumn
%\frenchspacing
%\setlength{\abovedisplayskip}{0pt}
%\setlength{\belowdisplayskip}{0pt}
\input{abstract}
\input{introduction_new6}

\input{related_work}
\input{overview_new3}
\input{experiments}
\input{conclusion}
%\input{discussion}
%\bibliographystyle{acm}
%\bibliography{atri_short}

%{\footnotesize
%{\small
	%\fontsize{9.0pt}{10.0pt}
	\small
	\bibliographystyle{aaai}
	\bibliography{atri_short}
%} 	
%}
%\newpage
%\onecolumn
%\input{supplementary}
%\input{conclusion}
\end{document}

%% file: title_and_authors.tex
\title{\textbf{Hypertree Decompositions Revisited for PGMs}}

\author {Aarthy Shivram Arun$^1$, Sai Vikneshwar Mani Jayarman$^1$, Christopher R\'e$^{2}$ and Atri Rudra$^1$
	\\
	$^1$ University at Buffalo, SUNY,
	\{ashivram, saivikne, atri\}@buffalo.edu.\\ 
	$^2$ Stanford University, 
	chrismre@cs.stanford.edu.\\
}

%% file: abstract.tex
\begin{abstract}
We revisit the classical problem of exact inference on probabilistic graphical models (PGMs). Our algorithm is based on recent \emph{worst-case optimal database join} algorithms, which can be asymptotically faster than traditional data processing methods. We present the first empirical evaluation of these algorithms via {\ourModel} -- a new exact inference engine. We empirically explore the properties of the data for which our engine can be expected to outperform traditional inference engines, refining current theoretical notions. Further, {\ourModel} outperforms existing state-of-the-art inference engines (ACE, IJGP and libDAI) on some standard benchmark datasets by up to a factor of 630x. Finally, we propose a promising data-driven heuristic that extends {\ourModel} to automatically tailor its parameters and/or switch to the traditional inference algorithms. 
\end{abstract}

%% file: introduction_new6.tex
\section{Introduction}
Efficient inference on probabilistic graphical models (PGMs) is a core topic in artificial intelligence (AI) and standard inference techniques are based on tree decompositions~\cite{jensen1990,dechter1996,kask2005,IJGP}. The runtime of such inference algorithms is exponential in the treewidth ($\tw$) of the underlying graph, which in the worst case, is unavoidable. Over the years, efforts in the logic, database and AI communities to refine $\tw$ into a finer-grained measure of complexity have culminated in {\em generalized hypertree decompositions} (GHDs)~\cite{fischl2016,gottlob2005}. Recently, FAQ/AJAR~\cite{faq,ajar} theoretically reconnected such GHD-based algorithms with probabilistic inference exploiting recent developments in \emph{worst-case optimal database join} algorithms~\cite{ngo} and achieved tighter bounds based on a finer-grained notion of width called {\em fractional-hypertreewidth} ($\fhtw$). Further, given the known connection between database joins and CSPs~\cite{csp-1,csp-2}, their bounds apply to many classes of CSPs as well.

However, the practical significance of GHD-based inference algorithms has met with some skepticism so far. In particular, Dechter et al.~\cite{dechter08} use a predictive ratio ($R$) (based on hypertreewidth $\htw$ and $\tw$) and concluded that classical treewidth-based algorithms outperform their GHD-based counterparts on an overwhelming majority of PGM benchmarks. Further, their study suggests that the advantages of GHD-based algorithms manifests only in instances with substantial factor sparsity (i.e. a large number of factor entries having zero probabilities) and high factor arity. 

In addition to the above constraints, translating the superior theoretical guarantees of GHD-based algorithms into practice is a non-trivial challenge. In particular, these algorithms~\cite{faq,ajar} typically assume that one can exhaustively search over all potential GHDs (which grows exponentially in the number of variables) and their runtime analysis ignores the dependence on number of variables/factors. Unfortunately, in practice, these assumptions could negate the theoretical gains.

\paragraph{Our Contributions.} In this paper, we introduce {\ourModel}, a proof-of-concept inference engine, to address all the above problems: {\ourModel} is efficient and can be up to $630$x faster than its competitors. Further, we introduce a theoretical measure that better predicts when {\ourModel} can outperform its competitors.

\textbf{\textit{GHDs Revisited:}} We empirically demonstrate that {\ourModel} (a GHD-based algorithm) has wider scope than previously predicted, revisiting the conclusions of~\cite{dechter08}. We do so with two new measures -- $\rho$, the total number of entries processed across all bags of the GHD and $R_{J}$, a better predictor of {\ourModel}'s performance. We observe that $R_D$ (our analogous version of $R$ from~\cite{dechter08} where $\htw$ is replaced by $\fhtw$), is contingent upon $\rho$ (see $\rho$ high and $\rho$ low classes in Figure~\ref{introTab}), an insight that is not captured in~\cite{dechter08}'s experimental paradigm. In particular, engines employing truth-table indices (libDAI/IJGP) do not scale to higher values of $\rho$, whereas {\ourModel} does (see Bands $1-3$ in Figure~\ref{introTab}). For instance, in \bandthree~in Figure~\ref{introTab}, {\ourModel} outperforms competing engines by up to $2.7$x, while $R_D$ predicts under-performance by more than $10^{10}$x. 

We then exploit the better theoretical runtime analysis of {\ourModel} to introduce $R_J$, a finer-grained theoretical measure that better predicts its performance. For instance, consider the `$R_D$ small' column in Figure~\ref{introTab} -- when $\rho$ is high (\bandone~in Figure~\ref{introTab}), {\ourModel} is up to $630$x faster than competing engines and when $\rho$ is low (\bandfour~in Figure~\ref{introTab}), {\ourModel} is only up to $5$x faster. We note that $R_J$ can actually differentiate between these two rows while $R_D$ predicts a similar speedup for both bands (see columns $4$-$7$ in Table~\ref{table_real_3}). 

\textbf{\textit{Technical Contributions:}} {\ourModel} leverages recently introduced worst-case optimal join algorithms~\cite{ngo2} in conjunction with improved data structures. In particular, we use two data representations for use in different passes of the algorithm -- a level-order trie, which collapses a conventional trie into a single array and (two variants of) an index-based compressed list. We find that the resulting gains more than compensate for the overheads involved in maintaining both data representations.

\textbf{\textit{Hybrid Architecture:}} Given the relative advantages of {\ourModel} (e.g, Bands $1-4$ in Figure~\ref{introTab}) and libDAI (e.g., Bands $5-6$ in Figure~\ref{introTab}) in different spaces, we explore the feasibility of exploiting their strengths in a `best-of-all-worlds' architecture. Our hybrid system (HJYAR) outperforms libDAI, IJGP and ACE for $75\%$ of the networks (see Table~\ref{table_real_3}), illustrating its promise.

\input{overview_fig}

%% file: overview_fig.tex
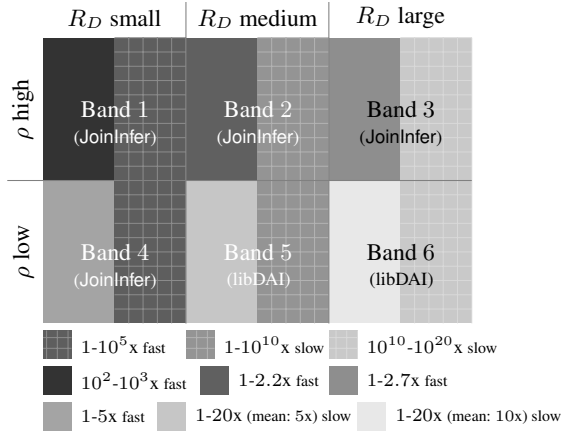
\begin{figure}[t]
\begin{center}
\begin{tikzpicture}[scale = 1.9]

%Vertical Label
\node[above, rotate=90] at (.5,1.5) {{\small {$\rho$ low}}};
\node[above, rotate=90] at (.5,2.5) {{\small {$\rho$ high}}};

%Horizontal Lables
\node[above] at (1,3) {{\small {$R_D$ small}}};
\node[above] at (2,3) {{\small {$R_D$ medium}}};
\node[above] at (3,3) {{\small {$R_D$ large}}};

%Legend for R_D
\fill[color=red!90] (0.5, 0.75) rectangle (0.7, 0.95);
\draw[step=0.75mm, gray!90, very thin] (.5,.75) grid (.7,.95);
\node[right] at (.7,.85) {{\scriptsize {$1$-$10^5$x \tiny fast}}};
\fill[color=red!60] (1.5, 0.75) rectangle (1.7, 0.95);
\draw[step=0.75mm, gray!60, very thin] (1.5,.75) grid (1.7,.95);
\node[right] at (1.7,.85) {{\scriptsize {$1$-$10^{10}$x \tiny slow}}};
\fill[color=red!30] (2.5, 0.75) rectangle (2.7, 0.95);
\draw[step=0.75mm, gray!30, very thin] (2.5,.75) grid (2.7,.95);
\node[right] at (2.7,.85) {{\scriptsize {$10^{10}$-$10^{20}$x \tiny slow}}};

%Legend for JI
\fill[color=blue!90] (.5,.5) rectangle (.7,.7);
\node[right] at (.7,.6) {{\scriptsize {$10^2$-$10^3$x \tiny fast}}};
\fill[color=blue!70] (1.6,.5) rectangle (1.8,.7);
\node[right] at (1.8,.6) {{\scriptsize {$1$-$2.2$x \tiny fast}}};
\fill[color=blue!50] (2.5,.5) rectangle (2.7,.7);
\node[right] at (2.7,.6) {{\scriptsize {$1$-$2.7$x \tiny fast}}};

\fill[color=blue!40] (0.5,.25) rectangle (0.7,.45);
\node[right] at (0.7,.35) {{\scriptsize {$1$-$5$x \tiny fast}}};
\fill[color=blue!25] (1.3,.25) rectangle (1.5,.45);
\node[right] at (1.5,.35) {{\scriptsize {$1$-$20$x \tiny (mean: $5$x) slow}}};
\fill[color=blue!10] (2.7,.25) rectangle (2.9,.45);
\node[right] at (2.9,.35) {{\scriptsize {$1$-$20$x \tiny (mean: $10$x) slow}}};

%Coloring of cells
%High rho
%band one
\fill[color=blue!90] (.5,2) rectangle (1,3);
\fill[color=red!90] (1,2) rectangle (1.5,3);
\draw[step=1mm,gray!90, very thin] (1,2) grid (1.5,3);
%band two
\fill[color=blue!70] (1.5,2) rectangle (2,3);
\fill[color=red!60] (2,2) rectangle (2.5,3);
\draw[step=1mm,gray!60, very thin] (2,2) grid (2.5,3);
%band three
\fill[color=blue!50] (2.5,2) rectangle (3,3);
\fill[color=red!30] (3,2) rectangle (3.5,3);
\draw[step=1mm,gray!30, very thin] (3,2) grid (3.5,3);
%Low rho
%band four
\fill[color=blue!40] (.5,1) rectangle (1,2);
\fill[color=red!90] (1,1) rectangle (1.5,2);
\draw[step=1mm,gray!90, very thin] (1,1) grid (1.5,2);
%band five
\fill[color=blue!25] (1.5,1) rectangle (2,2);
\fill[color=red!60] (2,1) rectangle (2.5,2);
\draw[step=1mm,gray!60, very thin] (2,1) grid (2.5,2);
%band six
\fill[color=blue!10] (2.5,1) rectangle (3,2);
\fill[color=red!30] (3,1) rectangle (3.5,2);
\draw[step=1mm,gray!30, very thin] (3,1) grid (3.5,2);

%Lables for cells
%Band name
\node[right] at (.7,2.5) {{\footnotesize \textcolor{white}{\bandone}}};
%Winner engine
\node[right] at (.65,2.3) {{\scriptsize \textcolor{white}{(\ourModel)}}};
%Band name
\node[right] at (1.7,2.5) {{\footnotesize \textcolor{white}{\bandtwo}}};
%Winner engine
\node[right] at (1.65,2.3) {{\scriptsize \textcolor{white}{(\ourModel)}}};
%Band name
\node[right] at (2.7,2.5) {{\footnotesize \textcolor{black}{\bandthree}}};
%Winner engine
\node[right] at (2.65,2.3) {{\scriptsize \textcolor{black}{(\ourModel)}}};
%Band name
\node[right] at (.7,1.5) {{\footnotesize \textcolor{white}{\bandfour}}};
%Winner engine
\node[right] at (.65,1.3) {{\scriptsize \textcolor{white}{(\ourModel)}}};
%Band name
\node[right] at (1.7,1.5) {{\footnotesize \textcolor{white}{\bandfive}}};
%Winner engine
\node[right] at (1.72,1.3) {{\scriptsize \textcolor{white}{(libDAI)}}};
%Band name
\node[right] at (2.7,1.5) {{\footnotesize \textcolor{black}{\bandsix}}};
%Winner engine
\node[right] at (2.72,1.3) {{\scriptsize \textcolor{black}{(libDAI)}}};

%Draw the cells
%Low \rho boxes
%\draw[color=gray] (.5,1) -- (1.5,1) -- (1.5,2) -- (.5,2) -- (.5,1);
%\draw[color=gray] (1.5,1) -- (2.5,1) -- (2.5,2) -- (1.5,2) -- (1.5,1);
%\draw[color=gray] (2.5,1) -- (3.5,1) -- (3.5,2) -- (2.5,2) -- (2.5,1);
%Low \rho high
%\draw[color=gray] (.5,2) -- (1.5,2) -- (1.5,3) -- (.5,3) -- (.5,2);
%\draw[color=gray] (1.5,2) -- (2.5,2) -- (2.5,3) -- (1.5,3) -- (1.5,2);
%\draw[color=gray] (2.5,2) -- (3.5,2) -- (3.5,3) -- (2.5,3) -- (2.5,2);

\draw[color=gray] (.25,2) -- (3.5,2);
\draw[color=gray] (1.5,3.25) -- (1.5,1);
\draw[color=gray] (2.5,3.25) -- (2.5,1);

\end{tikzpicture}
\end{center}
\caption{\small{Datasets are divided  into six bands depending on the sizes of $R_D$ and $\rho$. The grids in each box denote the expected speedup of any GHD-based system over a treewidth based system and gray shades show the actual speedup of \ourModel\ with respect to libDAI. The ``winner" is stated explicitly for each band.}}
%\caption{\small{Runtime Scope of {\ourModel}: `Small' denotes `$10^{-6} < R_D \leq 1$', 'Medium' denotes `$1 < R_D \leq 10^{3}$', `Large' denotes `$10^{3} < R_D \leq 10^{20}$'.`JI' categorizes where {\ourModel} is the best engine and `LD' categorizes where libDAI is the best engine.} \yell{I have tokens on this figure-- Atri} \yell{Check: \bandone\ \bandtwo\ \bandthree\ \bandfour\ \bandfive\ \bandsix\ }}
\label{introTab}
\end{figure}

%% file: related_work.tex
\section{Related Work} \label{sec:relWork}
Several streams of inquiry have emerged in the exact inference setting. One such stream involves \emph{conditioning algorithms}~\cite{pearl1989,darwicheRecursive} that adopt a case-based reasoning approach. Another class of algorithms seeks to exploit local structure~\cite{LarkinDechter03,pooleZhang03}, where~\cite{chavira_07,huang06,ijcai2005,darwicheRecursive} exploit factor sparsity to improve tractability of inference. Recently, an emerging area \emph{lifted probabilistic inference}~\cite{braz05,milch08,kersting12}, exploits symmetric structures within graphs to speed up inference. Yet another line of work runs along variable elimination~\cite{dechter1996,ZhangPoole1996} and tree decomposition-based routines~\cite{jensen1990,kask2005,IJGP}. Finally, past work in PGMs has also focused on approximate inference~\cite{koller2009,bekker2015,thinjt}; we believe that advancements in {\ourModel} could enhance their performance.

%% file: overview_new3.tex
\section{\ourModel: An Overview} \label{sec:overview}
We start by giving a brief overview of the background concepts and outline the \emph{worst-case optimal join} algorithm {\NPRR} that {\ourModel} uses for computing factor products. We then show how {\NPRR} fits in {\ourModel} -- our GHD-based Message Passing Algorithm. Finally, we talk about implementation challenges and our solutions, including a Hybrid Architecture.
\subsection{Preliminaries and Notation}
\begin{definition} \label{Definition:pgm}
A (discrete) \emph{\textbf{probabilistic graphical model}} can be defined by the triplet $\langle \mathcal{H}, D,\mathcal{K} \rangle$, where hypergraph $\mathcal{H} = (\mathcal{V},\mathcal{E})$ represents the underlying graphical structure (note $\mathcal{E}\subseteq 2^{\mathcal{V}}$). There are $n=|\mathcal{V}|$ discrete random variables on finite domains $D = \{D(U): U \in \mathcal{V}\}$ and $m = |\mathcal{E}|$ factors $\mathcal{K} = \{\phi_e\}_{e \in \mathcal{E}}$, where each factor $\phi_e$ is a mapping: $\phi_e:\prod_{U\in e}D(U)\rightarrow \mathbb{R}_+$.
\end{definition}
For instance, Figure~\ref{fig:pgm} in the full version is a hypergraph representing a PGM with variables $\mathcal{V} = \{A, B, C, D\}$, edges $\mathcal{E} = \{e {=} (A, B)$, $f {=} (A, C)$, $g {=} (B, C, D)\}$ and factors $\mathcal{K}=\{\phi_{e}(A,B)$, $\phi_{f}(A,C)$, $\phi_{g}(B, C, D)\}$.
\begin{definition}
\label{defn:sparsity}
For any factor $\phi$, the size of $\phi$ is its support size, i.e., the number of entries with non-zero probabilities. Storing only the non-zero entries (as well as their $\phi$ values) is called the \emph{listing representation} of $\phi$. \emph{Factor sparsity} is defined as $\frac{N}{\prod\limits_{U \in \phi} |D(U)|}$, where $N$ is the size of factor $\phi$.
\end{definition}
A typical inference task in PGMs is to compute the marginal estimates given by: $\forall F \subseteq \mathcal{V}$, $\mathbf{y} \in \prod_{U\in F} D(U)$,
\begin{equation}\label{eq:marginal}
\phi_F(\mathbf{y}) = \frac{1}{Z} \sum_{\mathbf{z} \in \prod\limits_{U \in \mathcal{V} \setminus F}D(U)} \prod_{S \in \mathcal{E}} \phi_S(\mathbf{\mathrm{x}}_S), 
\end{equation}
where $\boldsymbol{\mathrm{x}} = (\mathbf{y}, \mathbf{z})$, $\mathbf{\mathrm{x}}_S$ denotes the projection of $\mathbf{\mathrm{x}}$ onto the variables in $S$ and $Z$ is a normalization constant. Variable/Factor marginals are a special case of (\ref{eq:marginal}); $F = \{U\}$ for $U \in \mathcal{V}$ for variable marginals and $F \in \mathcal{E}$ for factor marginals. 

Exact inference in PGMs is usually performed by propagating on a \emph{\textbf{generalized hypertree decomposition (GHD)}} of the underlying hypergraph $\mathcal{H}$. 
\begin{definition} \label{Definition:ghd}
A GHD of $\mathcal{H}=(\mathcal{V},\mathcal{E})$ is defined by a triple $ \langle T,\chi,\lambda \rangle$, where $T=(V(T), E(T))$ is a tree, $\chi: V(T) \rightarrow 2^{\mathcal{V}}$ is a function associating a set of vertices $\chi(v) \subseteq \mathcal{V}$ to each node $v$ of $T$, and $\lambda: V(T) \rightarrow 2^{\mathcal{E}}$ is a function associating a set of hyperedges to each node $v$ of $T$ such that the following properties hold (i) for each $e \in \mathcal{E}$, there is a node $v \in V(T)$ such that $e \subseteq \chi(v)$ and $e \in \lambda(v)$; and (ii) for every $V' \subseteq \mathcal{V}$, the set $\{v \in V(T) | V' \subseteq \chi(v)\}$ is connected in $T$. We define \emph{treewidth} $\tw  = \max_{v \in V(T)}(|\chi{(v)}|)$.
\end{definition}

\textbf{\textit{Existing GHD-based Message Passing Algorithms.}} A GHD can be thought of as a labeled (hyper)tree $T$, where sets assigned to each node in $T$ are called \emph{bags} of the hypertree. Inference propagation on $T$ involves a two-pass \lq{}message-passing\rq{} algorithm~\cite{jensen1990}. In the first pass (message-up), \emph{messages} are propagated \lq{}up\rq{} from leaf (child) to root (parent). Subsequently in the second pass (message-down), they are propagated \lq{}down\rq{} from root (parent) to leaf (child).

More formally, a \emph{message} $\phi\rq{}_{m_{v,u}}$ from node $v$ to node $u$ is a marginal estimate given by $\sum_{U \notin \chi{(v)} \cap \chi{(u)}} \phi'_{v}$, where 
\begin{equation}\label{eq:joint}
\phi\rq{}_v = \prod_{e \in \lambda(v)} \phi_e \cdot \prod_{w \in \children(v)} \phi\rq{}_{m_{w,v}},
\end{equation}
for the upward pass. For the downward pass, $\children(v)$ will be replaced by $\parent{(v)}$. Upon completion of both passes (up/down), variable marginals for all $U \in \chi(v)$ can be retrieved using label $\chi(v)$ and factor marginals for all $e \in \lambda(v)$ can be retrieved using $\lambda{(v)}$ from each node $v \in V(T)$. Computing $\phi'_{v}$ is the major bottleneck in message-passing algorithms and we focus on this step next.

\subsection{\NPRR: A New Algorithm for Computing Factor Products}
In this section, we describe \NPRR, the \emph{worst-case optimal join} algorithm used by {\ourModel} to compute $\phi'_{v}$ for every $v \in V(T)$. We first present a motivating example, followed by the runtime analysis of {\NPRR}.
\paragraph{Triangle Query} \label{subsection:triangle}
As an example, consider the \emph{triangle} PGM with variables $\mathcal{V} = \{A, B, C\}$, edges $\mathcal{E} = \{e = (A, B), f = (B, C), g = (A, C)\}$ and factors $\mathcal{K} = \{\phi_{e}{(A, B)}, \phi_{f}{(B, C)}, \phi_{g}{(A, C)}\}$. Let $|D(U)| = D$ for all $U \in \mathcal{V}$ and $|\phi_{e}(A, B)| =  |\phi_{f}(B, C)| = |\phi_{g}(A, C)| = N \leq D^2$. We would like to compute the factor product $\phi'{(A, B, C)} = \phi_{e}(A, B) \cdot \phi_{f}(B, C) \cdot \phi_{g}(C, A)$ since the GHD for this PGM would contain only one node with variables $(A, B, C)$. Prior to~\cite{faq,ajar}, there are two algorithms used for computing factor products:
\begin{itemize}
 \item {The first would go over all $D^3$ possible output tuples $(a,b,c)$ and compute the product of corresponding probability values, resulting in an overall runtime of $O(D^3)$. This algorithm is called MultiplyFactors [\cite{darwiche}, Chapter $6$].}
 \item {The second would compute an intermediate product (say) $\phi_{e}(A,B) \cdot \phi_{f}(B,C)$ to get up to $N^2$ possible tuples $(a,b,c)$ and then filter this against $\phi_{g}(A,C)$. This takes $O(N^2)$ time and is faster than $O(D^3)$ if $N \ll D^{\frac{3}{2}}$.} 
\end{itemize}
We would like to mention here that the engines libDAI and IJGP compute intermediate pairwise products on a truth-table indexing scheme, staying closer to the runtime of $O(D^3)$ (see full version of this algorithm). The most interesting thing about \emph{worst-case join algorithms} (and hence {\NPRR})~is that they compute the above product in time $O(N^{\frac{3}{2}})$, which is asymptotically better than both the above algorithms as long as $N \ll D^{2}$. The key insight in {\NPRR} is that if $\phi_{e}(A,B) \cdot \phi_{f}(B,C) \cdot \phi_{g}(A,C)$ is computed in a multiway fashion, then one can exploit sparsity in the input factors (which MultiplyFactors fails to do) and avoid computing larger intermediate products (which \PP~fails to do). As an extreme example, consider the case when $\phi_{e}(A,B)$ has $N$ non-zero tuples of the form $[N]\times[1]$ and $\phi_{f}(B,C)$ has $N$ non-zero tuples of the form $[1] \times [N]$. The pairwise product $\phi_{e}(A,B) \cdot \phi_{f}(B,C)$ will thus have $N^2$ non-zero entries $[N] \times [1] \times [N]$. However, we know only $N$ of them would survive since $\phi_{g}(A,C)$ has exactly $N$ non-zero entries. In this case, {\NPRR} will first fix variable $A$'s value to $a$ and $B$'s value of $1$ (i.e, $(a, 1) \in \phi_{e}(A, B)$) and then obtain all values of $c$ such that $(1, c) \in \phi_{f}(B, C)$ and $(a, c) \in \phi_{g}(A, C)$. Note that there exists a unique value of $c$ for a fixed $a$. Thus, {\NPRR} will process only $N$ entries overall in this case since $|\phi_{e}(A, B)| \le N$. (See Algorithm~\ref{nprr_s} for our algorithm for the triangle case and full version for general algorithm.)

\input{nprr_simplified}

\subsubsection{Runtime Complexity of {\NPRR}} \label{sec:runtime}
A recent result of Atserias, Grohe and Marx showed how to tightly bound the worst-case output size of a factor product~\cite{agm} and subsequently,~\cite{ngo} came up with {\NPRR} that can run in time of the worst-case output size. For a hypergraph $\mathcal{H} = (\mathcal{V},\mathcal{E})$, let $\mathbf{x} \in \mathbb{R}^{|\mathcal{E}|}$ be a vector indexed by edges, such that $\mathbf{x}^* = \left(x_e\right)_{e \in \mathcal{E}}$ is an optimal solution to the linear program
\begin{eqnarray}\label{eq:lpAGM}
\min && \sum_{e \in \mathcal{E}} x_e \text{log}_2 |\phi_e|\\ 
\text{s.t.} && \sum_{v \ni e} x_e \geq 1 \space \forall v \in \mathcal{V}; x_e \geq 0 \forall e \in \mathcal{E}.
\end{eqnarray}
Then, we can bound the size of the factor product $\phi' = \prod_{e \in \mathcal{E}} \phi_{e}$ as follows:
\begin{equation} \label{eq:AGMbound}
|\phi'| \le \prod_{e \in \mathcal{E}} |\phi_e|^{x_e}.
\end{equation}
In particular, the runtime of {\NPRR} is $O(\prod_{e \in \mathcal{E}} |\phi_e|^{x_e})$ and is reflected in the triangle example we considered earlier.
\begin{example}
In the previous section, we considered the triangle query with edges $e = (A, B)$, $f = (B, C)$ and $g = (A, C)$ respectively. Solving the linear program~\eqref{eq:lpAGM} for this case, we have $\mathbf{x}^{*} = (\frac{1}{2}, \frac{1}{2}, \frac{1}{2})$. Note that this gives us an asymptotically better bound of $N^{\frac{3}{2}}$ since $|\phi_{e}(A, B)| = |\phi_{f}(B, C)| = |\phi_{g}(A, C)| = N$. In comparison, the hypertreewidth-based bound is $N^{2}$ $(\htw{(T)} = 2)$ and the treewidth-based bound is $D^{3}$ $(\tw{(T)} = 3)$.
\end{example}

\subsection{Using {\NPRR} in GHD-based Message Passing Algorithms}
The GHD-based Message Passing Algorithm (also known as Junction-Tree Algorithm) forms the structure of {\ourModel} (we present a sketch of this procedure in Algorithm~\ref{jta}). The input PGM network is transformed into a Junction Tree (i.e., a GHD) using the min-fill heuristic. Then, we root the GHD arbitrarily, determining the parent-child relationship for every node. We assume that each input factor is assigned to a unique bag in the GHD. In particular, for every node $v \in V(T)$, $\alpha{(v)}$ denotes the input factors assigned to it.
\begin{algorithm} [H]\label{junction_tree_algorithm}
	\caption{\ourModel} \label{jta}
	\begin{algorithmic}[1]
		\small
		\State{\textbf{Input:} A PGM $\mathcal{P} = (\mathcal{H}, D, \mathcal{K})$.}
		\State{\textbf{Output:} Variable and Factor Marginals.}
		\State{\text{Create a rooted GHD $\mathcal{G} = ((V,E), \chi, \lambda)$ for $\mathcal{P}$.}} \Comment{Using min-fill variable ordering.}
		\label{line:JI-ghd}
		\State $R\gets \ourModelSampling(\mathcal{G})$ \Comment{$R : V \to \{0, 1, 2\}$ and $\ourModelSampling$ (i.e., Algorithm in the full version.}
		\State{$(\{\phi'_{v}\}_{v \in V}, \{\phi'_{m_{v, \parent(v)}}\}_{v \in V})\gets \ourModelUp(\mathcal{G} , R)$}
		\State{$\{\phi'_{v}\}_{v \in V}\gets$}
		\Statex{\quad \quad $\ourModelDown(\mathcal{G},  \{\phi'_{v}\}_{v \in V},\{\phi'_{m_{v, \parent(v)}}\}_{v \in V})$}
		\State{\text{Compute Variable and Factor Marginals from} $(\{\phi'_{v}\}_{v \in V})$}
	\end{algorithmic}
\end{algorithm}
We now describe our contributions in the message-up and message-down phases.

\subsubsection{Message-Up Phase}
The upward pass propagates messages (i.e., marginalized factor products) from leaves to root along the rooted GHD. Recall that this involves computing factor products as in~\eqref{eq:joint} at every node (bag) of the GHD. {\ourModel} lends its core contribution in this phase -- it uses {\NPRR} to compute the factor products. As shown in the Triangle example earlier, {\NPRR} is different from previously proposed exact inference algorithms and performs a multiway product to achieve asymptotically better bounds (see Algorithm~\ref{nprr_s} for an outline). We present the general algorithm in the full version (adapted from~\cite{ngo}).

Further, for a given bag, in addition to input factors mapped/messages received by it, {\ourModel} includes factors that are not originally mapped to the bag, but have non-trivial intersections with variables in the bag (using ideas from Section $3.2$ in~\cite{dechter08} and we call these factors ``01-projections"). Note that these are crucial for theoretical results for FAQ/AJAR and can prune factor product entries early on in the presence of factor sparsity. 

We present our Message-Up procedure in Algorithm~\ref{up}.\footnote{When $R_{v} = 2$, {\PP} (Algorithm in the full version) is run instead of {\NPRR}.}
\begin{algorithm} [H] \label{message_up}
	\caption{\ourModelUp} \label{up}
	\begin{algorithmic}[1]
		\small
		\State{\textbf{Input:} GHD $\mathcal{G} = ((V,E), \chi, \lambda)$ and the map $R$.}%i.e. the output of Algorithm~\ref{sampAlgo}.}
		\State{\textbf{Output:} Factor Products $\{\phi'_{v}\}_{v \in V}$ and Up Messages $\{\phi'_{m_{v, \parent{(v)}}}\}_{v \in V}$ both as tries.}
		\ForAll{\text{nodes} $v \in V$}: \Comment{This is done in a level-order traversal from leaves-to-root.}
		\State{$\mathcal{E}_{v}\gets \lambda(v), \mathcal{K}_{v} \gets \{\phi_{e}: e \in \lambda(v)\}$}  \Comment{Initialize the PGM Query corresponding to $v$'s Factor Product.}
		\ForAll{$w \in \children{(v)}$} \Comment{We add all the messages sent to $v$ from its children.}
		\State{$\mathcal{E}_{v} \gets \mathcal{E}_{v} \cup \{\chi{(v)} \cap \chi{(w)}\}, \mathcal{K}_{v} \gets \mathcal{K}_{v} \cup \{\phi'_{m_{w, v}}\}$}
		\EndFor
		\If{$R(v) = 1$} \Comment{We include the $0/1$ projections while computing the Factor Product for $v$. } \label{line:JI-up-chk-R}
		\ForAll{$e \in \mathcal{E} \setminus \lambda{(v)}$} 
		\If{$e \cap \chi{(v)} \neq \emptyset$} 
		\State{$\mathcal{E}_{v} \gets \mathcal{E}_{v} \cup \{e \cap \chi{(v)}\}, \mathcal{K}_{v} \gets \mathcal{K}_{v} \cup \{\phi_{e / \chi{(v)}}\}$}
		\EndIf
		\EndFor
		\EndIf
		\If{$v$\text{ is not a root}}
		\State{Let $u = \parent(v)$, $\phi'_{m_{v, u}}$ be $\sum_{q \in \chi{(v)} \setminus \chi{(u)}} \phi'_{v}$ and $\mathcal{F}_{v} = \chi{(v)} \cap \chi{(u)}$.}
		\State{$(\phi'_{v}, \phi'_{m_{v, u}}) \gets \NPRR(R_{v}, \chi{(v)}, \mathcal{E}_{v}, \mathcal{K}_{v}, \mathcal{F}_{v})$} 
		\label{line:call-nprr-1}
		\Else{ $\phi'_{v} \gets \NPRR(R_{v}, \chi{(v)}, \mathcal{E}_{v}, \mathcal{K}_{v}, \chi{(v)})$}
		\label{line:call-nprr-2}
		\EndIf
		\EndFor
	\end{algorithmic}
\end{algorithm}

\subsubsection{Message-Down Phase}
The downward pass (from root-to-leaves) involves updating factor products for each bag (except the root) using the received messages. Note that there are two products computed for each bag -- one between the sent and received message and the second, between the result of the previous step with the bag's factor product. We use \emph{in-place hash products} for this computation (more details in the full version).

\subsubsection{Runtime Complexity: An Analysis of {\ourModel}}
For a GHD $G = ( T = (V, E), \chi, \lambda)$, we can bound the size of $\phi'_{u}$ for every $u \in V(T)$ as follows. For a hypergraph $\mathcal{H}_{u} = (\mathcal{V}_{\chi(u)},\mathcal{E})$, let $\mathbf{x} \in \mathbb{R}^{|\mathcal{E}|}$ be a vector indexed by edges, such that $\mathbf{x}^*_{u} = \left(x^{u}_e\right)_{e \in \mathcal{E}}$ is an optimal solution to \eqref{eq:AGMbound}, bounding $|\phi'_{u}|$:
\begin{equation} \label{eq:AGMbound2}
|\phi'_{u}| \le \prod_{e \in \mathcal{E}} |\phi_e|^{x^{u}_e}.
\end{equation}
Since we run {\NPRR} for every $u \in V(T)$, the total runtime of {\ourModel} is $O(\sum_{u \in V(T)} \prod_{e \in \mathcal{E}} |\phi_e|^{x^{u}_e})$.

Upper-bounding each $|\phi_e|$ by $N = \max_{e \in \mathcal{E}} |\phi_{e}|$ in the above equation and replacing the sum over all $u \in V(T)$ by max gives us an asymptotic bound of $N^{\fhtw(T)}$. $\fhtw(T)$ is guaranteed to be at most $\htw(T)$ (hypetreewidth), which in turn is at most $\tw(T)$ (treewidth) for the same GHD, giving us the best known theoretical bounds for exact inference in PGMs. (More details in the full version.)

Recall that the asymptotic bounds for $\tw$ based algorithms is given by $D^{\tw}$, where $D = \max_{U \in \mathcal{V}} |D(U)|$. However, a more realistic measure here would be $\rho = \sum_{u \in V(T)} \prod_{U \in \chi(u)}| D(U)|$. This gives us a fine-grained ratio (as compared to~\cite{dechter08}) to evaluate {\ourModel} against classical engines:
\begin{equation}\label{eq:eqRJ}
R_J = \frac{\sum_{u \in V(T)} \prod_{e \in \mathcal{E}} |\phi_e|^{x^{u}_{e}}}{\rho}.
\end{equation}
Replacing the numerator with $N^{\fhtw}$ and the denominator with $D^{\tw}$ in \eqref{eq:eqRJ} gives us
\begin{equation}\label{eq:eqRD}
R_D = \frac{N^{\fhtw}}{D^{\tw}}.
\end{equation}
This ratio is analogous to the one in~\cite{dechter08}, which was based on hypertree width:
\[ R = \log_{10}\left(\frac{N^{\htw^*}}{D^{tw}}\right).\]
Since computing $\htw$ is NP-Hard,~\cite{dechter08} used an approximation for it (denoted by $\htw^*$). In our measure $R_{D}$, we overcome this issue by using $\fhtw$ over $\htw$. Using $\fhtw$ offers two significant advantages -- one, it is a more fine-grained measure (since $\fhtw \le \htw \le \tw$) and two, $\fhtw$ is polynomially computable (basically, solve the LP from~\eqref{eq:lpAGM}). We empirically demonstrate that $R_J$ is a better predictor than $R_D$ in Section~$4.2.1$ since it is a more fine-grained measure.

\subsection{Technical Contributions}
Consistent with \cite{faq,ajar}, we represent every factor table as a trie (storing only entries with non-zero probabilities),\footnote{A trie is a multi-level data structure where each factor tuple corresponds to a unique path from root to leaf and the probability value associated with each tuple are stored in the leaf.} in our implementation of {\NPRR} and use a \emph{flattened} version of the tries to exploit caching advantages. In addition to this, we store factor tables as lists of $\langle \text{index value, probability} \rangle$ pairs, where each factor tuple is converted into a corresponding index value. We use two variants -- the first stores only `reverse' indices (computed in reverse variable order) and the second stores forward and reverse indices (for representing intermediary messages). Note that these representations enable us to optimize the up and down passes. In particular, the reverse index enables efficient construction of tries in the up-phase and in decoding message entries over all children in a single pass in the down-phase. Moreover, the reverse indices of the up-messages act as placeholders for down-messages, enabling the reuse of data structures. Finally, the forward indices are used while merging down-messages with cluster products in-place, thus optimizing decoding/encoding steps.

\subsection{Hybrid Architecture}
In Bands $5-6$ (see Figure~\ref{introTab}), libDAI's pairwise product implementation demonstrates distinct advantages over {\ourModel}'s multi-way product. We explore the feasibility of leveraging the respective advantages of both these strategies in a new \{HY\}brid \{J\}oin \{AR\}chitecture (HYJAR). To build such a system, we use the native structure of {\ourModel} and import only the pairwise-product functionality from libDAI (we do not integrate the entire engine). Given the high costs of switching between the data structures required for {\ourModel} and libDAI, the main challenge here was to devise a system that not only optimally chooses between the strategies per bag, but at the same time minimizes the switches between bags.\footnote{We would like to mention here that ACE uses a similar rule at the GHD-level but ours works at the bag-level.} We overcome this challenge by introducing a deterministic heuristic that decides the optimal strategy ({\ourModel} or {\PP}) for each bag $v$ in the GHD that has at least one input factor assigned to it (i.e. $\alpha(v)\ge 1$).\footnote{Recall our earlier assumption that each factor table is assigned to a unique bag. As a result, not many bags are chosen in this process. Further we ignore the incoming messages for a bag $v$ when deciding on $R_{v}$, making this decision faster.} We then propagate this decision along the subtree of $v$, until it reaches a bag that was already assigned a decision. To decide the order of preference, we consider bags $v$ in decreasing order of $\prod_{U\in \chi(v)}|D(U)|$, with the intuition that larger bags dominate the runtime of libDAI. More details in the full version.
%We present this Algorithm~\ref{hybridAlgo} and its description in the full version.

%% file: nprr_simplified.tex
\begin{algorithm} [H] \label{mult_fac_prod}
	\caption{{\NPRR} for Triangle} \label{nprr_s}
	\begin{algorithmic}[1]
		\small
		\State{\textbf{Input:} Variables $\mathcal{V} =\{A, B, C\}$, Edges $\mathcal{E} = \{e = (A, B), f = (B, C), g = (A, C)\}$ and Factors $\mathcal{K} = \{ \phi_{e}(A, B), \phi_{f}(B, C), \phi_{g}(A, C) \}$.} 
		\State{\textbf{Output:} Factor Product $\phi'(A, B, C) = \phi_{e}(A, B) \cdot \phi_{f}(B, C) \cdot \phi_{g}(A, C)$.} 
		\ForAll{$a \text{ s.t. } \exists b, c \text{ with } \phi_{e}(A = a, B = b) \cdot \phi_{f}(A = a, C = c) \neq 0$}
			\ForAll{$b \text{ s.t. } \exists c \text{ with } \phi_{e}(A = a, B = b) \cdot \phi_{f}(B = b, C = c) \neq 0$} \Comment{Value for variable $A$ is fixed as $a$.}
				\ForAll{$c \text{ s.t. } \phi_{e}(A = a, B = b) \cdot \phi_{f}(B = b, C = c) \neq 0$} \Comment{Value for variables $A$ and $B$ are fixed as $a$ and $b$ respectively.}
					\State{$\phi'(A, B, C) \gets \phi'(A, B, C) \cup \{ (a, b, c), \phi_{e}{(a, b)} \cdot \phi_{f}{(b, c) \cdot \phi_{g}{(a, c)}}\}$} \Comment{The entry $(a, b, c)$ is added to the factor product along with its corresponding probability.}
				\EndFor
			\EndFor
		\EndFor
\State{\Return{$\phi'(A, B, C)$}}
\end{algorithmic}
\end{algorithm}

%% file: experiments.tex
\section{Experimental Evaluation}
In this section, we empirically validate {\ourModel} and outline features that influence its performance. We start by describing our empirical setup introducing the standard benchmarks and demonstrate the scope of {\ourModel} vis-a-vis state-of-the-art-systems on them. Then, we document performance gains of our hybrid system and finally, evaluate our technical contributions.

\subsection{Experimental Setup}\label{sec:exp_setup}
We first create a testbed of $52$ networks that spans the full range of cases illustrated in Figure~\ref{introTab}, sampling from three publicly available benchmarks -- UAI\rq{}06~\cite{uai2006}, PIC 2011~\cite{pascal11} and the BN Learns dataset~\cite{bnLearn} (which subsumes IJCAI\rq{}05 networks~\cite{ijcai2005}). Further, in order to improve the tractability of some of the larger networks (Bands $1$, $2$ and $3$) for exact inference (high $\rho$ cases), we randomly induce factor sparsity.\footnote{Details in the full version.} Note that these sparsity levels are consistent with the ranges found in other networks in the benchmark and inducing sparsity to improve model tractability is a well-accepted procedure in many practical settings~\cite{songHanICLR18,LarkinDechter03}. For the {\bandtwo} networks BN\_$30-39$, we modify the original probabilities (these networks are marked with a `*' in Table~\ref{table_real_3}) and provide details in the full version.

We compare {\ourModel} against three state-of-the-art systems on the exact inference query of computing all variable marginals: ACE~\cite{ijcai2005}, an engine that explicitly exploits determinism, and, libDAI~\cite{libDAI} and IJGP~\cite{IJGP}, two award winning systems in the UAI 2010 inference challenge. Additionally, while {\ourModel}, IJGP and ACE process evidence, libDAI does not. Hence, to ensure a fair comparison, we incorporate evidence information directly into the input given to the engines,\footnote{Details in the full version.} We compare our marginal outputs with these engines, with an error limit of $0.00001$. We evaluate all the systems on the average time taken across $5$ runs to compute all the variable marginals (setting a timeout of $60$ minutes for each run). Further, ACE requires separate compilation of the arithmetic circuit representing the input network (a non-standard design). For a fair comparison with other engines with end-to-end computations, we report two times for ACE -- the sum of compilation and inference times, followed by inference time. 

We ran all our experiments on a Linux server (Ubuntu $16.04$ LTS) with Intel Xeon E$5$-$2640$ v$3$ CPU @ $2.60$GHz and $64$ GB RAM. 
\subsection{Experimental Results}\label{sec:exp}
\subsubsection{Benchmark Experiments} \input{headline_table_new}
The results in Table~\ref{table_real_3} are laid out along the lines of Figure~\ref{introTab}. These networks span over a wide range of sparsity ($20\% - 100\%$), domain sizes ($2 -100$) and factor arity levels ($1-10$).
	
\textbf{\textit{$\rho$ High.}} The measure $R_D$ from~\cite{dechter08} predicts superior performance for {\ourModel} only in {\bandone} (CELAR).  However, in this region of high $\rho$, {\ourModel} performs consistently better than the predictions in~\cite{dechter08}. In {\bandone}, it is be up to $630$x faster on subsets where ACE completes. In {\bandtwo}, it is up to $2.2$x faster and in {\bandthree} where the corresponding predictions of~\cite{dechter08} is under-performance by $10^{10}\times$ - $10^{20}\times$, it can be upto $2.7$x faster than ACE (libDAI and IJGP fail in this space). libDAI and IJGP fail in these bands due to huge pre-memory allocation. On the other hand, ACE that takes advantage of {\em factor sparsity} using arithmetic circuits is the only other engine that completes; that said, compiling these structures is costly. In \bandone, we surmise that {\ourModel}'s performance advantages are rooted in the use of {\NPRR} as opposed to standard algorithms, which is predicted by $R_J$. The networks in these bands ($1-3$) cover a sparsity range of $20\%-50\%$ and have factor arity level $1-4$. Further, in \bandone, given that ACE requires the $20\%$ sparsity levels to complete, we present results at two levels of sparsity for CELAR ($20\%$ and $40\%$).

\textbf{\textit{$\rho$ Low.}} $R_D$ predicts superior performance for {\ourModel}  in {\bandfour}, which it achieves. It is upto $5.29$x faster than libDAI (it's closest competitor) and up to $5.4$x faster than ACE (on the subsets that ACE completes on). IJGP times out on almost all of the networks. Finally, in Bands $5/6$, the two unfavorable settings, {\ourModel} is on an average faster than ACE by $5.8$x/$9.5$x and IJGP by $2.36$x/$2.28$x respectively. It is on an average slower than libDAI by $4.8\times$/$10\times$ respectively: libDAI's truth-table indexing advantages clearly manifest in these two bands. We would like to note however, that the corresponding predictions of ~\cite{dechter08} for {\ourModel} is under performance by $10$x - $10^8$x for {\bandfive}, and $10^{20}$x - $10^{28}$x for {\bandsix}, i.e., several orders of magnitude worse.
	
\textbf{\textit{Predictions by $R_J$.}} Our finer grained ratio $R_J$ (Column $4$) better tracks {\ourModel}'s speed-ups as compared to $R_D$ (Column $5$) on most networks. Though it is similar to $R_D$ in {\bandone} and within couple of orders magnitude in {\bandtwo}, it diverges considerably (more than $10^5$x) in Bands $3$ and $4$ (where {\ourModel}'s speed-ups are within $1$x - $5$x). Moreover, $R_J$ has much tighter predictions than $R_D$ across almost all of the networks in the testbed.

In addition to the above, we find that {\ourModel} is competitive in terms of memory usage and the use of ``01-projections" do not help significantly in our benchmarks (more details in the full version).

\subsubsection{Hybrid Architecture (HYJAR).} Since {\ourModel} is the only engine that completes on all networks when $\rho$ is high, we now focus on low $\rho$ conditions. HYJAR helps exploit the relative strengths of each strategy--multiway or pairwise products--into a single architecture, yielding consistent performance across a majority of networks ($26/28$). Of these, in $9$ cases (e.g., munin1, munin, barley, mildew) HYJAR's completion times are faster than its nearest standalone competitors ({\ourModel} or libDAI), in $10$ cases it is less than $2.5$x slower and in $7$ cases it is between $2.5$x - $4.5$x slower. BN\_$42$ and BN\_$44$ are the only two networks where HYJAR's strategy does not lead to a notable improvement. Further, follow up analysis indicates that HYJAR consistently switches between strategies at the bag level (more details in the full version).
	
\subsubsection{Factor Representations.} As described in the Overview Section, we store two variants of the list of $\langle \text{index, probability} \rangle$ pairs -- one with a forward index and the other with a reverse index. Our experiments on the UAI speech recognition datasets show that this results in up to $3$x, $1.6$x and $1.3$x gains during building tries, message-up and message-down phases respectively (details in the full version). 

\subsubsection{Takeaways.} We have identified a threshold for $\rho$ at $10^9$ reflecting the current memory limits for truth tables (on our machine).\footnote{While the absolute value of the threshold may change depending on machine configurations, such a threshold will always exist.} Further, we have demonstrated that $R_J$ is a better predictor of {\ourModel}'s performance than $R_D$. Finally, HYJAR outperforms libDAI, IJGP and ACE on $39$ out of $52$ networks (i.e., $75\%$ of them), illustrating its promise as a practically relevant architecture for building a robust, broadly applicable inference engine. Note that the second largest winner is libDAI with wins on only $11$ datasets.

%% file: headline_table_new.tex
\begin{table*}
	\centering
	\caption{{ \small Benchmark Comparisons: The first column denotes the range of $\rho$, followed by the bands of the datasets (see Figure~\ref{introTab}) and the dataset name. The fourth column denotes the number of variables/factors, followed by $R_{J}$ and $R_{D}$. We report three runtimes for {\ourModel}: without 0/1 projections, with all 0/1 projections and HYJAR, followed by our comparison engines -- LibDAI, IJGP and ACE (Total Time and Inference Time). (All runtimes are in seconds.) Further, we report the median and mean sparsity for every dataset, followed by fractional hypertree width ($\fhtw$) and tree-width ($\tw$) (computed for the same GHD). The fractional hypertreewidth ($\fhtw$) numbers were generated by solving the linear program~\eqref{eq:lpAGM} using Google OR-Tools~\cite{orsolver}. Finally, we report the maximum domain value (D) and maximum factor table (non-zero) entry size (N). \lq{}T\rq{} denotes engine-time out (60 mins). The networks BN\_$30-39$ are denoted by `*' since we modify the probabilities in them (details in the full version). Note that LibDAI and IJGP crash on all benchmarks where $\rho > 10^{9}$, due to huge pre-memory allocation and this is denoted by \lq{}F\rq{}. For IJGP, we observed that it does approximate inference in benchmarks Munin1 and BN$\_43-46$ respectively. In particular, we recorded its final treewidth using the MinFill ordering on all benchmarks and compared it with {\ourModel}'s and libDAI's treewidth (both using the MinFill ordering) respectively. We noticed that the final treewidth reported by IJGP was much smaller than the treewidth reported by {\ourModel} and libDAI. Note that we preprocess evidence and SAT-based singleton consistency on all the benchmarks and thus, we concluded that IJGP does Approximate Inference on these datasets (which we denote by \lq{}A\rq{}).}}
	\label{table_real_3}
	\resizebox{1.0\textwidth}{!}{
	\begin{tabular}{|c|cccccccccccccccc|}

	$\rho$ & Band & Dataset & Var/Factors & $R_{J}$ & $R_{D}$ & \multicolumn{2}{c}{\ourModel}  & HYJAR & libDAI & IJGP & \multicolumn{2}{c}{ACE}  & Sparsity (in \%) & fhtw & tw & D/N \\
	\hline
	& & & & & & w/o 0/1 & 0/1 & & & & TTime & ITime & & & & \\
	\toprule
	\multirow{24}{*}{$\rho > 10^{9}$} & \multirow{8}{*}{\bandone} & CELAR6-SUB0\_20 & 16/57 & 2.00E-03 & 1.00E-03 & 0.17 & \textbf{0.16} & 0.19 & F & F & 97.24 & 0.36 & 20/20 & 4 & 8 & 44/387 \\

	& & CELAR6-SUB1\_20 & 14/75 & 3.00E-04 & 3.00E-04 & \textbf{2.57} & 2.58 & 2.59 & F & F & 444.38 & 0.99 & 20/20 & 5 & 10 & 44/387 \\

	& & CELAR6-SUB2\_20 & 16/89 & 1.00E-04 & 1.00E-04 & 1.05 & \textbf{1.04} & 1.07 & F & F & 653.33 & 0.74 & 20/20 & 5.5 & 11 & 44/387 \\

	& & CELAR6-SUB3\_20 & 18/106 & 1.00E-04 & 1.00E-04 & 3.72 & 3.69 & \textbf{3.67} & F & F & 1219.08 & 0.78 & 20/20 & 5.5 & 11 & 44/387 \\

	& & CELAR6-SUB0\_40 & 16/57 & 3.00E-02 & 2.50E-02 & 4.55 & 4.59 & \textbf{4.17} & F & F & 855.12 & 1.2 & 40/40 & 4 & 8 & 44/774 \\

	& & CELAR6-SUB1\_40 & 14/75 & 1.00E-02 & 1.00E-02 & 392.7 & 388.76 & \textbf{388.42} & F & F & T & T & 40/40 & 5 & 10 & 44/774 \\

	& & CELAR6-SUB2\_40 & 16/89 & 6.50E-03 & 6.40E-03 & 449.02 & 448.56 & \textbf{441.41} & F & F & T & T & 40/40 & 5.5 & 11 & 44/774 \\

	& & CELAR6-SUB3\_40 & 18/106 & 6.00E-03 & 7.00E-03 & 796.76 & 794.98 & \textbf{780.93} & F & F & T & T & 40/40 & 5.5 & 11 & 44/774 \\
	
	\cline{2-17}

	& \multirow{11}{*}{\bandtwo} & BN\_30* & 1036/1153 & 15.96 & 5.10E+02 & 1.05 & 1.11 & \textbf{1.01} & F & F & 2.20 & 0.28 & 50/44.5 & 25 & 41 & 2/4 \\

	& & BN\_31* & 1036/1153 & 47.47 & 2.00E+03 & 1.04 & 1.1 & \textbf{1.03} & F & F & 2.23 & 0.31 & 50/44.5 & 25 & 39 & 2/4 \\

	& & BN\_32* & 1294/1441 & 1.88 & 1.20E+02 & 1.61 & 1.71 & \textbf{1.60} & F & F & 2.94 & 0.32 & 50/44.3 & 28 & 49 & 2/4 \\

	& & BN\_33* & 1294/1441 & 6.20E+02 & 1.00E+03 & 1.62 & 1.68 & \textbf{1.60} & F & F & 2.86 & 0.31 & 50/44.3 & 26 & 42 & 2/4 \\

	& & BN\_34* & 1294/1443 & 2.60E+04 & 3.20E+04 & 1.60 & 1.66 & \textbf{1.59} & F & F & 2.76 & 0.31 & 50/44.3 & 28 & 41 & 2/4 \\

	& & BN\_35* & 1294/1443 & 3.10E+01 & 5.10E+02 & 1.59 & 1.65 & \textbf{1.58} & F & F & 2.82 & 0.31 & 50/44.3 & 26 & 43 & 2/4 \\

	& & BN\_36* & 1294/1444 & 1.70E+03 & 2.00E+03 & 1.62 & 1.70 & \textbf{1.60} & F & F & 3.05 & 0.31 & 50/43.7 & 30 & 49 & 2/4 \\

	& & BN\_37* & 1294/1444 & 7.10E+02 & 1.00E+03 & 1.64 & 1.70 & \textbf{1.62} & F & F & 2.76 & 0.30 & 50/43.7 & 30 & 50 & 2/4 \\

	& & BN\_38* & 1294/1442 & 4.10E+02 & 2.50E+02 & 1.58 & 1.65 & \textbf{1.55} & F & F & 2.94 & 0.32 & 50/43.9 & 27 & 46 & 2/4 \\

	& & BN\_39* & 1294/1442 & 10.7 & 2.50E+02 & 1.61 & 1.65 & \textbf{1.60} & F & F & 2.93 & 0.32 & 50/43.9 & 26 & 44 & 2/4 \\

	& & BN\_62 & 657/667 & 7.70E+09 & 8.38E+09 & 0.74 & 0.7 & \textbf{0.68} & F & F & 1.45 & 0.24 & 25/34.1 & 21 & 47 & 2/14 \\
	
	\cline{2-17}
	
	& \multirow{5}{*}{\bandthree} & BN\_60 & 530/539 & 1.20E+08 & 7.20E+16 & \textbf{0.7} & 0.75 & 0.73 & F & F & 1.65 & 0.24 & 50/44.03 & 29 & 60 & 2/16 \\
	
	&  & BN\_61 & 657/667 & 3.00E+10 & 1.70E+10 & 0.74 & 0.74 & \textbf{0.69} & F & F & 1.69 & 0.24 & 25/34.1 & 21 & 46 & 2/14 \\
	
	& & BN\_63 & 530/540 & 1.40E+10 & 9.20E+18 & 2.43 & 0.77 &\textbf{0.67} & F & F & 1.84 & 0.27 & 50/43.5 & 30 & 57 & 2/16 \\

	& & BN\_64 & 530/540 & 1.60E+09 & 5.76E+17 & 0.79 & 0.68 & \textbf{0.63} & F & F & 1.84 & 0.25 & 50/43.5 & 28.5 & 55 & 2/16 \\

	& & BN\_67 & 430/437 & 4.60E+10 & 2.95E+20 & 2.64 & \textbf{1.65} & 2.05 & F & F & 1.71 & 0.26 & 50/50.57 & 32.5 & 62 & 2/16 \\
	\midrule
	\multirow{28}{*}{$\rho \le 10^{9}$} & \multirow{7}{*}{\bandfour} & BN\_20 & 2433/2840 & 119 & 1.00E-04 & 4.53 & \textbf{4.47} & 14.94 & 22.73 & T & T & T & 50/49.3 & 4 & 7 & 91/208 \\

	& & BN\_21 & 2433/2840 & 109 & 1.00E-04 & 4.49 & \textbf{4.42} & 14.86 & 23.37 & T & T & T & 50/49.3 & 4 & 7 & 91/208 \\

	& & BN\_22 & 2119/2423 & 0.97 & 1.00E-05 & \textbf{2.13} & 2.18 & 2.98 & 3.77 & T & 7.83 & 1.71 & 50/47 & 4 & 7 & 91/208 \\

	& & BN\_23 & 2119/2423 & 0.97 & 1.00E-05 & \textbf{2.14} & 2.21 & 3 & 3.74 & T & 7.74 & 1.79 & 50/47 & 2 & 5 & 91/208 \\

	& & BN\_24 & 1514/1818 & 2.08 & 1.00E-05 & \textbf{1.33} & 1.4 & 1.74 & 2.11 & T & 6.24 & 1.58 & 53.8/53 & 2 & 5 & 91/208 \\

	& & BN\_25 & 1514/1818 & 2.01 & 1.00E-05 & \textbf{1.31} & 1.39 & 1.76 & 2.12 & T & 6.34 & 1.62 & 53.8/53 & 2 & 5 & 91/208 \\

	& & Pathfinder & 109/109 & 54.94 & 1.00E-05 & 0.15 & 0.29 & 0.29 & \textbf{0.11} & 0.34 & 0.81 & 0.31 & 52.4/61.4 & 2 & 7 & 63/6437 \\
	
	\cline{2-17}
	
	& \multirow{15}{*}{\bandfive} & Alarm & 37/37 & 3.58 & 11.39 & 0.03 & 0.03 & 0.05 & \textbf{0.02} & 0.06 & 0.46 & 0.21 & 100/99.4 & 2 & 5 & 4/108 \\

	& & Hepar2 & 70/70 & 5.95 & 9 & 0.04 & 0.05 & 0.06 & \textbf{0.03} & 0.19 & 0.42 & 0.19 & 100/100 & 2 & 7 & 4/384 \\

	& & Mildew & 35/35 & 2.00E+03 & 327 & 0.78 & 0.71 & \textbf{0.24} & 0.27 & 2.81 & 2.91 & 1.89 & 75/61.7 & 3 & 5 & 100/14849 \\

	& & Munin & 1041/1041 & 35.77 & 557 & 13.27 & 13.82 & \textbf{1.98} & 3.14 & 11.35 & T & T & 42.1/46.6 & 6 & 9 & 21/276 \\

	& & Munin1 & 186/186 & 43.23 & 16.6 & 598.86 & 629.75 & \textbf{20.6} & 39.01 & A & T & T & 46.2/48.6 & 7 & 12 & 21/276 \\

	& & Munin4 & 1038/1038 & 57.9 & 557 & 16.86 & 17.21 & 2.15 & \textbf{2.06} & 10.67 & 3.77 & 2.01 & 44/46.6 & 6 & 9 & 21/276 \\

	& & Diabetes & 413/413 & 524.51 & 4.24E+06 & 3.28 & 3.31 & \textbf{0.72} & 0.89 & 32.98 & 6.99 & 4.69 & 33.3/45.6 & 4 & 5 & 21/2040 \\

	& & Munin2 & 1003/1003 & 2.90E+05 & 8.90E+08 & 2.71 & 2.28 & \textbf{0.68} & 0.79 & 4.27 & 2.81 & 1.63 & 46.4/48 & 8 & 8 & 21/276 \\

	& & Munin3 & 1041/1041 & 5.00E+05 & 1.20E+04 & 2.71 & 2.52 & \textbf{0.70} & 0.95 & 5.81 & 2.38 & 1.28 & 45.8/37 & 6 & 8 & 21/276 \\

	& & Pigs & 441/441 & 144 & 1.50E+04 & 0.98 & 0.92 & 0.36 & \textbf{0.24} & 1.05 & 1.37 & 0.7 & 55.6/70.2 & 8 & 11 & 3/15 \\

	& & Link & 724/724 & 2.00E+07 & 2.70E+08 & 18.02 & 19.91 & 14.27 & \textbf{3.43} & 29.73 & 201.02 & 7.49 & 50/65.1 & 12 & 16 & 4/31 \\

	& & Barley & 48/48 & 4.00E+07 & 6.50E+03 & 26.69 & 27.17 & \textbf{1.13} & 1.45 & 15.32 & 17.53 & 10.94 & 100/100 & 4 & 8 & 67/40320 \\

	& & Hailfinder & 56/56 & 13.02 & 1.00E+04 & 0.03 & 0.05 & 0.06 & \textbf{0.01} & 0.05 & 0.53 & 0.23 & 94.2/83.9 & 3 & 5 & 11/1181 \\

	& & Water & 32/32 & 1.00E+05 & 1.00E+06 & 0.17 & \textbf{0.14} & 0.27 & 0.31 & 0.15 & 0.81 & 0.34 & 50/58.23 & 4 & 11 & 4/1454 \\

  & & Win95pts & 76/76 & 8.66 & 3.10E+04 & 0.05 & 0.05 & 0.05 & \textbf{0.03} & \textbf{0.03} & 0.53 & 0.2 & 100/90 & 3 & 9 & 2/252 \\
	\cline{2-17}
	& \multirow{6}{*}{\bandsix} & Andes & 223/223 & 3.10E+04 & 1.50E+20 & 0.57 & 0.59 & 0.19 & \textbf{0.14} & 0.59 & 1.12 & 0.67 & 100/95.7 & 12 & 17 & 2/128 \\

	& & BN\_42 & 870/879 & 1.23 & 4.72E+21 & 32.63 & 32.11 & 35.15 & \textbf{2.66} & 19.18 & 1216.39 & 19.6 & 50/54.4 & 24 & 24 & 2/16 \\

	& & BN\_43 & 870/880 & 1.14 & 3.78E+22 & 65.47 & 64.03 & \textbf{4.37} & 4.43 & A & 1132.1 & 22.24 & 50/54.4 & 25 & 25 & 2/16 \\

	& & BN\_44 & 870/880 & 1.03 & 2.40E+24 & 227.87 & 216.98 & 133.5 & \textbf{12.82} & A & 1341.72 & 17.02 & 50/54.3 & 27 & 27 & 2/16 \\

	& & BN\_45 & 870/880 & 1.1 & 3.78E+22 & 67.05 & 68.21 & 8.07 & \textbf{6.95} & A & 778.91 & 19.05 & 50/54.2 & 25 & 25 & 2/16 \\

	& & BN\_46 & 489/497 & 1.04 & 7.92E+28 & 45.98 & 46.09 & 20.16 & \textbf{5.85} & A & 150.25 & 5.79 & 50/55.9 & 24 & 24 & 2/16 \\

	%& & C432.isc & 432/432 & 4.40E+27 & 7.60E+53 & 45.98 & 46.01 & 34.3 & 31.12 & 105.08 & 141.46 & 2.41 & 50/54.17 & 23 & 28 & 2/512 \\

	%& & C499.isc & 499/499 & 1.04 & 1.27E+30 & 7.52 & 7.49 & 13.25 & 15.77 & A & 11.65 & 0.82 & 50/54.11 & 25 & 25 & 2/32 \\

	%& & C880.isc & 880/880 & 1.16 & 3.78E+22 & 0.53 & 0.53 & 2.43 & 4.91 & A & 1.10 & 0.23 & 50/53.41 & 25 & 25 & 2/16 \\
	\bottomrule
\end{tabular}}
\end{table*}

%% file: conclusion.tex
\paragraph{Conclusion} Our system can be extended to classes of CSPs studied in~\cite{faq}, which we think are potentially relevant for inference on Lifted Graphical Models.

%% file: ms.bbl
\begin{thebibliography}{}

\bibitem[\protect\citeauthoryear{Abo~Khamis, Ngo, and Rudra}{2016}]{faq}
Abo~Khamis, M.; Ngo, H.~Q.; and Rudra, A.
\newblock 2016.
\newblock {FAQ}: questions asked frequently.
\newblock In {\em Proc. 35th PODS},  13--28.
\newblock ACM.

\bibitem[\protect\citeauthoryear{Atserias, Grohe, and Marx}{2013}]{agm}
Atserias, A.; Grohe, M.; and Marx, D.
\newblock 2013.
\newblock Size bounds and query plans for relational joins.
\newblock {\em {SIAM} J. Comput.} 42(4):1737--1767.

\bibitem[\protect\citeauthoryear{Bach and Jordan}{2001}]{thinjt}
Bach, F.~R., and Jordan, M.~I.
\newblock 2001.
\newblock Thin junction trees.
\newblock In {\em Advances in Neural Information Processing Systems},
  569--576.

\bibitem[\protect\citeauthoryear{Bekker \bgroup et al\mbox.\egroup
  }{2015}]{bekker2015}
Bekker, J.; Davis, J.; Choi, A.; Darwiche, A.; and Van~den Broeck, G.
\newblock 2015.
\newblock Tractable learning for complex probability queries.
\newblock In {\em Advances in Neural Information Processing Systems},
  2242--2250.

\bibitem[\protect\citeauthoryear{Bilmes and Dechter}{2006}]{uai2006}
Bilmes, J., and Dechter, R.
\newblock 2006.
\newblock Evaluation of probabilistic inference systems of {UAI}’06.

\bibitem[\protect\citeauthoryear{bnLearn}{}]{bnLearn}
bnLearn.
\newblock Bayesian network repository.
\newblock \url{http://bnlearn.com/bnrepository/}.

\bibitem[\protect\citeauthoryear{Cai, Lu, and Xia}{2014}]{csp-2}
Cai, J.; Lu, P.; and Xia, M.
\newblock 2014.
\newblock The complexity of complex weighted boolean {\#}csp.
\newblock {\em J. Comput. Syst. Sci.} 80(1):217--236.

\bibitem[\protect\citeauthoryear{Chavira and Darwiche}{2005}]{ijcai2005}
Chavira, M., and Darwiche, A.
\newblock 2005.
\newblock Compiling bayesian networks with local structure.
\newblock In {\em IJCAI}, volume~5,  1306--1312.

\bibitem[\protect\citeauthoryear{Chavira and Darwiche}{2007}]{chavira_07}
Chavira, M., and Darwiche, A.
\newblock 2007.
\newblock Compiling bayesian networks using variable elimination.
\newblock In {\em Proc. 20th IJCAI},  2443--2449.

\bibitem[\protect\citeauthoryear{Chen and Grohe}{2010}]{csp-1}
Chen, H., and Grohe, M.
\newblock 2010.
\newblock Constraint satisfaction with succinctly specified relations.
\newblock {\em J. Comput. Syst. Sci.} 76(8):847--860.

\bibitem[\protect\citeauthoryear{Darwiche}{2001}]{darwicheRecursive}
Darwiche, A.
\newblock 2001.
\newblock Recursive conditioning.
\newblock {\em Artif. Intell.} 126(1-2):5--41.

\bibitem[\protect\citeauthoryear{Darwiche}{2009}]{darwiche}
Darwiche, A.
\newblock 2009.
\newblock {\em Modeling and Reasoning with Bayesian Networks}.
\newblock Cambridge University Press.

\bibitem[\protect\citeauthoryear{de Salvo~Braz, Amir, and Roth}{2005}]{braz05}
de~Salvo~Braz, R.; Amir, E.; and Roth, D.
\newblock 2005.
\newblock Lifted first-order probabilistic inference.
\newblock In {\em Proc. 19th IJCAI},  1319--1325.

\bibitem[\protect\citeauthoryear{Dechter, Otten, and
  Marinescu}{2008}]{dechter08}
Dechter, R.; Otten, L.; and Marinescu, R.
\newblock 2008.
\newblock On the practical significance of hypertree vs. treewidth.
\newblock In {\em {ECAI}}, volume 178,  913--914.

\bibitem[\protect\citeauthoryear{Dechter}{1996}]{dechter1996}
Dechter, R.
\newblock 1996.
\newblock Bucket elimination: {A} unifying framework for probabilistic
  inference.
\newblock In {\em Proc. 12th UAI},  211--219.

\bibitem[\protect\citeauthoryear{Fischl, Gottlob, and
  Pichler}{2016}]{fischl2016}
Fischl, W.; Gottlob, G.; and Pichler, R.
\newblock 2016.
\newblock General and fractional hypertree decompositions: Hard and easy cases.
\newblock {\em arXiv preprint arXiv:1611.01090}.

\bibitem[\protect\citeauthoryear{Gal~Elidan}{}]{pascal11}
Gal~Elidan, A.~G.
\newblock {Probabilistic Inference Challenge} 2011.
\newblock \url{http://www.cs.huji.ac.il/project/PASCAL/}.

\bibitem[\protect\citeauthoryear{Google}{}]{orsolver}
Google.
\newblock {OR Solver}.
\newblock \url{https://developers.google.com/optimization/}.

\bibitem[\protect\citeauthoryear{Gottlob \bgroup et al\mbox.\egroup
  }{2005}]{gottlob2005}
Gottlob, G.; Grohe, M.; Musliu, N.; Samer, M.; and Scarcello, F.
\newblock 2005.
\newblock Hypertree decompositions: Structure, algorithms, and applications.
\newblock In {\em WG (5) 2005},  1--15.
\newblock Springer.

\bibitem[\protect\citeauthoryear{Huang, Chavira, and Darwiche}{2006}]{huang06}
Huang, J.; Chavira, M.; and Darwiche, A.
\newblock 2006.
\newblock Solving {MAP} exactly by searching on compiled arithmetic circuits.
\newblock In {\em AAAI}, volume~6,  1143--1148.

\bibitem[\protect\citeauthoryear{Jensen, Lauritzen, and
  Olesen}{1990}]{jensen1990}
Jensen, F.; Lauritzen, S.; and Olesen, K.
\newblock 1990.
\newblock Bayesian updating in causal probabilistic networks by local
  computations.
\newblock {\em Computational Statistics Quarterly} 4:269--282.

\bibitem[\protect\citeauthoryear{Joglekar, Puttagunta, and
  R{\'{e}}}{2016}]{ajar}
Joglekar, M.~R.; Puttagunta, R.; and R{\'{e}}, C.
\newblock 2016.
\newblock {AJAR:} aggregations and joins over annotated relations.
\newblock In {\em Proc. 35th PODS},  91--106.

\bibitem[\protect\citeauthoryear{Kask \bgroup et al\mbox.\egroup
  }{2005}]{kask2005}
Kask, K.; Dechter, R.; Larrosa, J.; and Dechter, A.
\newblock 2005.
\newblock Unifying tree decompositions for reasoning in graphical models.
\newblock {\em Artificial Intelligence} 166(1-2):165--193.

\bibitem[\protect\citeauthoryear{Kersting}{2012}]{kersting12}
Kersting, K.
\newblock 2012.
\newblock Lifted probabilistic inference.
\newblock In {\em Proc. 20th {ECAI}},  33--38.

\bibitem[\protect\citeauthoryear{Koller and Friedman}{2009}]{koller2009}
Koller, D., and Friedman, N.
\newblock 2009.
\newblock {\em Probabilistic graphical models: principles and techniques}.
\newblock MIT press.

\bibitem[\protect\citeauthoryear{Larkin and Dechter}{2003}]{LarkinDechter03}
Larkin, D., and Dechter, R.
\newblock 2003.
\newblock Bayesian inference in the presence of determinism.
\newblock In {\em Proc. 9th AISTATS}.

\bibitem[\protect\citeauthoryear{Liu \bgroup et al\mbox.\egroup
  }{2018}]{songHanICLR18}
Liu, X.; Pool, J.; Han, S.; and Dally, W.~J.
\newblock 2018.
\newblock Efficient sparse-winograd convolutional neural networks.
\newblock {\em CoRR} abs/1802.06367.

\bibitem[\protect\citeauthoryear{Mateescu \bgroup et al\mbox.\egroup
  }{2010}]{IJGP}
Mateescu, R.; Kask, K.; Gogate, V.; and Dechter, R.
\newblock 2010.
\newblock Join-graph propagation algorithms.
\newblock {\em J. Artif. Intell. Res. {(JAIR)}} 37:279--328.

\bibitem[\protect\citeauthoryear{Milch \bgroup et al\mbox.\egroup
  }{2008}]{milch08}
Milch, B.; Zettlemoyer, L.~S.; Kersting, K.; Haimes, M.; and Kaelbling, L.~P.
\newblock 2008.
\newblock Lifted probabilistic inference with counting formulas.
\newblock In {\em Proc. 23rd {AAAI}},  1062--1068.

\bibitem[\protect\citeauthoryear{Mooij}{2010}]{libDAI}
Mooij, J.~M.
\newblock 2010.
\newblock lib{DAI}: A free and open source {C++} library for discrete
  approximate inference in graphical models.
\newblock {\em Journal of Machine Learning Research} 11:2169--2173.

\bibitem[\protect\citeauthoryear{Ngo \bgroup et al\mbox.\egroup }{2012}]{ngo}
Ngo, H.~Q.; Porat, E.; R{\'e}, C.; and Rudra, A.
\newblock 2012.
\newblock Worst-case optimal join algorithms:[extended abstract].
\newblock In {\em Proc. 31st PODS},  37--48.
\newblock ACM.

\bibitem[\protect\citeauthoryear{Ngo, R{\'e}, and Rudra}{2014}]{ngo2}
Ngo, H.~Q.; R{\'e}, C.; and Rudra, A.
\newblock 2014.
\newblock Skew strikes back: New developments in the theory of join algorithms.
\newblock {\em ACM SIGMOD Record} 42(4):5--16.

\bibitem[\protect\citeauthoryear{Pearl}{1989}]{pearl1989}
Pearl, J.
\newblock 1989.
\newblock {\em Probabilistic reasoning in intelligent systems - networks of
  plausible inference}.
\newblock Morgan Kaufmann series in representation and reasoning. Morgan
  Kaufmann.

\bibitem[\protect\citeauthoryear{Poole and Zhang}{2003}]{pooleZhang03}
Poole, D.~L., and Zhang, N.~L.
\newblock 2003.
\newblock Exploiting contextual independence in probabilistic inference.
\newblock {\em J. Artif. Intell. Res. {(JAIR)}} 18:263--313.

\bibitem[\protect\citeauthoryear{Zhang and Poole}{1996}]{ZhangPoole1996}
Zhang, N.~L., and Poole, D.~L.
\newblock 1996.
\newblock Exploiting causal independence in bayesian network inference.
\newblock {\em J. Artif. Intell. Res. {(JAIR)}} 5:301--328.

\end{thebibliography}
